%% file: main.tex
\documentclass[11pt,a4paper]{article}
\usepackage[hyperref]{acl2021}
\usepackage{times}
\usepackage{latexsym}

\usepackage{hyperref}
\usepackage{microtype}
\usepackage[utf8]{inputenc}
\usepackage{graphicx}
\usepackage{amsmath}
\usepackage{amsthm}
\usepackage{booktabs}
\usepackage{algorithm}
\usepackage{algorithmic}
\usepackage{amssymb,amsfonts}
\usepackage{textcomp}
\usepackage{xcolor}
\usepackage{todonotes}
\usepackage{booktabs,adjustbox,hhline}
\usepackage{multirow}
\usepackage{xspace}

\newcommand{\Ucal}{\mathcal{U}}
\newcommand{\Lcal}{\mathcal{L}}
\newcommand{\Xcal}{\mathcal{X}}
\newcommand{\Ycal}{\mathcal{Y}}
\newcommand{\Scal}{\mathcal{S}}
\newcommand{\bfl}{\mathbf l}
\newcommand{\bfx}{\mathbf x}

\newcommand{\model}{\mbox{\textsc{Spear}}}
\newcommand{\modelss}{\mbox{\textsc{Spear-SS}}}

\begin{document}
\aclfinalcopy

\title{Semi-Supervised Data Programming with Subset Selection}
\author{Ayush Maheshwari \textsuperscript{1}, Oishik Chatterjee \textsuperscript{1}, Krishnateja Killamsetty \textsuperscript{2}, \\ \textbf{Ganesh Ramakrishnan \textsuperscript{1} , Rishabh Iyer \textsuperscript{2} } \\
        \textsuperscript{1}Department of CSE, IIT Bombay, India \\ \textsuperscript{2} The University of Texas at Dallas  \\
\texttt{\{ayusham, oishik, ganesh\}@cse.iitb.ac.in}\\
\texttt{\{krishnateja.killamsetty, rishabh.iyer\}@utdallas.edu}
}

\maketitle

\begin{abstract}
The paradigm of data programming, which uses weak supervision in the form of rules/labelling functions, and semi-supervised learning, which augments small amounts of labelled data with a large unlabelled dataset, have shown great promise in several text classification scenarios. 
In this work, we argue that by not using any labelled data, data programming based approaches can yield sub-optimal performances, particularly when the labelling functions are noisy. The first contribution of this work is an introduction of a framework, \model \xspace which is a semi-supervised data programming paradigm that learns a \emph{joint model} that effectively uses the rules/labelling functions along with semi-supervised loss functions on the feature space. 
Next, we also study \modelss \xspace which additionally does subset selection on top of the joint semi-supervised data programming objective and \emph{selects} a set of examples that can be used as the labelled set by \model. The goal of \modelss \xspace is to ensure that the labelled data can \emph{complement} the labelling functions, thereby benefiting from both data-programming as well as appropriately selected data for human labelling. We demonstrate that by effectively combining semi-supervision,  data-programming, and subset selection paradigms, we significantly outperform the current state-of-the-art on seven publicly available datasets. \footnote{The source code is available at \url{https://github.com/ayushbits/Semi-Supervised-LFs-Subset-Selection}} 
\end{abstract}


\section{Introduction}
Modern machine learning techniques rely on large amounts of labelled training data for text classification tasks such as spam detection, (movie) genre classification, sequence labelling, {\em etc}. Supervised learning approaches have utilised such large amounts of labelled data and, this has resulted in huge successes in the last decade. However, the acquisition of labelled data, in most cases, entails a painstaking process requiring months of human effort. Several techniques such as active learning, distant supervision, crowd-consensus learning, and semi-supervised learning have been proposed to reduce the \textit{annotation cost}~\cite{annotation}. However, clean annotated labels continue to be critical for reliable results~\cite{bach2019snorkel,goh2018using}.
\par
Recently, \citet{ratner} proposed a paradigm on data-programming in which several Labelling Functions (LF) written by humans are used to weakly associate labels with the instances. In data programming, users encode the weak supervision in the form of labelling functions. On the other hand, traditional semi-supervised learning methods combine a small amount of labelled data with large unlabelled data~\cite{semisupervised}. 
In this paper, we leverage semi-supervision in the feature space for more effective data programming using labelling functions.

\subsection{Motivating  Example \label{sec:motivating}}
We illustrate the LFs on one of the seven tasks on which we experiment with, {\em viz.}, identifying spam/no-spam comments in the YouTube reviews. For some applications, writing LFs is often as simple as using keyword lookups or a regex expression. In this specific case, the users construct heuristic patterns as LFs for classifying spam/not-spam comments. Each LF takes a comment as an input and provides a binary label as the output; +1 indicates that the comment is spam, -1 indicates that the comment is not spam, and 0 indicates that the LF is unable to assert anything for the comment (referred to as an \textit{abstain}). Table~\ref{tab:sample} presents a few example LFs for spam and non-spam classification. 
\begin{table}[t]
\scriptsize
\begin{tabular}{|p{0.6cm}|p{6.3cm}|} 

\hline
{\bf Id} &  \textbf{Description} \\
\hline
{\bf LF1} &  If \texttt{http} or \texttt{https} in comment text, then return +1 otherwise ABSTAIN (return 0)\\
\hline
{\bf LF2} & If length of comment is less than 5 words, then return -1 otherwise ABSTAIN (return 0).(Non spam comments are often short)\\ \hline
{\bf LF3} & If comment contains \texttt{my channel} or \texttt{my video}, then return +1 otherwise ABSTAIN (return 0). \\ \hline
\end{tabular}
\caption{Three LFs based on keyword lookups or regex expression for the \textit{YouTube} spam classification task}\label{tab:sample}
\end{table}

In isolation, a particular LF may neither be always correct nor complete. Furthermore, the LFs may also produce conflicting labels.
In the past, generative models such as Snorkel~\cite{ratner} and CAGE~\cite{oishik} have been proposed for consensus on the noisy and conflicting labels assigned by the discrete LFs to determine the probability of the correct labels. Labels thus obtained could be used for training any supervised model/classifier and evaluated on a  test set. 
We will next highlight a challenge in doing data programming using only LFs that we attempt to address. 
For each of the following sentences $S_1 \ldots S_6$ that can constitute {\em an observed set of training instances},  we state the value of the true label ($\pm 1$). While the candidates in $S_1$ and $S_4$ are instances of a spam comment, the ones in $S_2$ and $S_3$ are not. In fact, these examples constitute one of the canonical cases that we discovered during the analysis of our approach in Section~\ref{sec:subsetselection}.\\
\noindent  1. $\langle \boldsymbol{S_1},+1 \rangle$: Please help me go to college guys! Thanks from the bottom of my heart.  
{\small https://www.indiegogo.com/projects/} \\ \normalsize 
\noindent 2. $\langle \boldsymbol{S_2},-1\rangle$: I love this song\\
\noindent 3. $\langle \boldsymbol{S_3},-1\rangle$: This song is very good... but the video makes no sense... \\
\noindent  4. $\langle \boldsymbol{S_4},+1\rangle$:  {\small https://www.facebook.com/teeLaLaLa } \\ 
Further, let us say we have a completely {\em unseen set of  test instances}, $S_5$ and $S_6$, whose labels we would also like to predict effectively: \\
\noindent 5. $\langle \boldsymbol{S_5},-1\rangle$: This song is prehistoric\\
\noindent 6. $\langle \boldsymbol{S_6},+1\rangle$:
Watch Maroon 5's latest ... {\small www.youtube.com/watch?v=TQ046FuAu00 }\normalsize


\begin{table}
\centering
\begin{tabular}{|c|c|c|c|c|c|} 
 \hline
 \multicolumn{2}{|c|}{Training data} & \multicolumn{2}{|c|}{LF outputs} & \multicolumn{2}{|c|}{Features} \\ [0.5ex] 
    id & Label & LF1(+1) & LF2(-1) & F1  & F2  \\ [0.5ex] 
 \hline
$S_1$ & +1 & 1 & 0 & 1 & 0  \\ 
$S_2$ & -1 & 0 & 1 & 0 &  1  \\ 
$S_3$ & -1 & 0 & 0 & 0 & 1  \\ 
$S_4$ & +1 & 1 & 1 & 1 & 0  \\ 
 \hline \hline
  \multicolumn{2}{|c|}{Test data} & \multicolumn{2}{|c|}{}  &  \multicolumn{2}{|c|}{} \\ \hline
$S_5$ &  -1 & 0 & 1 & 0 & 1  \\ 
$S_6$ & +1 & 0 & 0 & 1 &  0  \\ 
\hline
\end{tabular}
\caption{Example illustrating the insufficiency of using data programming using only LFs. \label{tab:insufficiency}}
\label{tab:twochallenges}
\end{table}
In Table~\ref{tab:twochallenges}, we present the outputs of the LFs as well as some  n-gram
features F1 (`.com') and F2  (`This song') on the observed training examples $S_1$,  $S_2$, $S_3$ and $S_4$ as well as on the unseen test examples $S_5$ and $S_6$. For $S_1$, the correct consensus can easily be performed to output the true label +1, since LF1 (designed for class +1) gets triggered, whereas LF2 (designed for class -1) is not triggered. Similarly, for $S_2$, LF2 gets triggered whereas LF1 is not, making it possible to easily perform the correct consensus. Hence, we have treated $S_1$ and $S_2$ as unlabelled, indicating that we could learn a model based on LFs alone without supervision if all we observed were these two examples and the outputs of LF1 and LF2. 
However, the correct consensus on $S_3$ and $S_4$  is challenging since both LF1 and LF2 either fire or do not. 
While the (n-gram based) features F1 and F2 appear to be informative and could potentially complement LF1 and LF2, we can easily see that correlating feature values with LF outputs is tricky in a completely unsupervised setup. To address this issue, we ask the following questions: \\
\noindent {\bf (A)} What if we are provided access to the true labels of a small subset of instances - in this case, only $S_3$ and $S_4$? Could the (i) correlation of features values ({\em eg.} F1 and F2) with labels ({\em eg.} +1 and -1 respectively), modelled via a small set of labelled instances ({\em eg.} $S_3$ and $S_4$), in conjunction with (ii) the correlation of feature values ({\em eg.} F1 and F2) with LFs ({\em eg.} LF1 and LF2) modelled via a potentially larger set of unlabelled instances ({\em eg.} $S_1$, $S_2$), help improved prediction of labels for hitherto unseen test instances $S_5$ and $S_6$? \\
\noindent {\bf (B)}  Can we precisely determine the subset of the unlabelled data that, when labelling would help us train a model (in conjunction with the labelling functions) that is most effective on the test set? In other words, instead of randomly choosing the labelled dataset for doing semi-supervised learning (part A), can we intelligently select the labelled subset?
In the above example, choosing the labelled set as $S_3, S_4$ would be much more useful than choosing the labelled set as $S_1, S_2$.


As a solution to (A), in Section~\ref{sec:spear}, we present a new formulation, \model,  in which the parameters over features and LFs are jointly trained in a semi-supervised manner. \model\ expands as {\bf S}emi-su{\bf P}ervis{\bf E}d d{\bf A}ta p{\bf R}ogramming. As for (B), we present a subset selection recipe, \modelss\ (in Section~\ref{sec:spearss}), that recommends the  sub-set of the data ({\em e.g.} $S_3$ and $S_4$), which, after labelling, would most benefit the joint learning framework.

\subsection{Our Contributions}
We summarise our main contributions as follows: To address (A), we present \model\ ({\em c.f.}, Section~\ref{sec:spear}), which is a novel paradigm for jointly learning the parameters over features and labelling functions in a semi-supervised manner. We jointly learn a parameterized graphical model and a classifier model to learn our overall objective. To address (B), we present \modelss\ ({\em c.f.}, Section~\ref{sec:spearss}), which is a subset selection approach to \emph{select} the set of examples which can be used as the labelled set by \model. We show, in particular, that through a principled data selection approach, we can achieve significantly higher accuracies than just randomly selecting the seed labelled set for semi-supervised learning with labelling functions. Moreover, we also show that the automatically selected subset performs comparably or even better than the hand-picked subset by humans in the work reported by  ~\citet{awasthi2020learning}, further emphasising the benefit of subset selection for semi-supervised data programming. 
Our framework is agnostic to the underlying network architecture and can be applied using different underlying techniques without a change in the meta-approach. Finally, we evaluate our model on seven publicly available datasets from domains such as spam detection, record classification, and genre prediction and demonstrate significant improvement over state-of-the-art techniques. We also draw insights from experiments in synthetic settings (presented in the appendix). 



\input{2.relatedwork_contributions}
\input{3.methodology}

\input{4.experiments}
\input{5.conclusion}
\section*{Acknowledgements}
We thank anonymous reviewers for providing constructive feedback. Ayush Maheshwari is supported by a Fellowship from Ekal Foundation (www.ekal.org). We are also grateful to IBM Research, India (specifically the IBM AI Horizon Networks - IIT Bombay initiative) for their support and sponsorship.
\bibliographystyle{acl_natbib}
\bibliography{ref.bib}
\appendix
\input{appendix}
\end{document}

%% file: 2.relatedwork_contributions.tex
\section{Related Work}
{\bf Data Programming and Unsupervised Learning:}  Snorkel~\cite{ratner} has been proposed as a generative model to determine correct label probability using consensus on the noisy and conflicting labels assigned by the discrete LFs. \citet{oishik} proposed a graphical model, CAGE, that uses continuous-valued LFs with scores obtained using soft match techniques such as cosine similarity of word vectors, TF-IDF score, distance among entity pairs, \textit{etc}. Owing  to its generative model, Snorkel is highly sensitive to initialisation and hyper-parameters. On the other hand, the CAGE model employs user-controlled quality guides that incorporate labeller intuition into the model. However, these models completely disregard feature information that could provide additional information to learn the (graphical) model. These models try to learn a combined model for the labelling functions in an unsupervised manner. However, in practical scenarios, some labelled data is always available (or could be made available by labelling a few instances); hence, a completely unsupervised approach might not be the best solution. In this work, we augment these data programming approaches by designing a  semi-supervised model that incorporates feature information and LFs to learn the parameters jointly. \citet{Hu2016HarnessingDN} proposed a student-teacher model that transfers rule information by assigning linear weight to each rule based on an agreement objective. The model we propose in this paper jointly learns parameters over features and rules in a semi-supervised manner rather than just weighing their outputs and can therefore be more expressive. 

Other works such as \citet{jawanpuria2011efficient, dembczynski2008maximum} discovers simple and conjuctive rules from input features and assign weight to each rules  for better generalization. \citet{nagesh-etal-2012-towards} induces rules from query language to build NER extractor while \citet{aaai2018} uses active learning to derive consensus among labelers to label data.

\noindent {\bf Semi-Supervised Data Programming:} The only work which, to our knowledge, combines rules with supervised learning in a joint framework is the work by \citet{awasthi2020learning}. They leverage both rules and labelled data by associating each rule with exemplars of correct firings ({\em i.e.}, instantiations) of that rule. Their joint training algorithms denoise over-generalized rules and train a classification model. Our approach differs from their work in two ways: a) we do not have information of rule exemplars - thus our labelled examples need not have any correspondence to any of the LFs (and may instead complement the LFs as illustrated in Table~\ref{tab:insufficiency}) and b) we employ a semi-supervised framework combined with graphical model for consensus amongst the LFs to train our model. We also study how to automatically select the seed set of labelled data, rather than having a human provide this seed set, as was done in~\cite{awasthi2020learning}.\looseness-1 

\noindent {\bf Data Subset Selection:} Finally, another approach that has been gaining a lot of attention recently is data subset selection. The specific application of data subset selection depends on the goal at hand. Data subset selection techniques have been used to reduce end to end training time~\cite{mirzasoleiman2019data,kaushal2019learning, killamsetty2020glister} and to select unlabelled points in an active learning manner to label~\cite{wei2015submodularity,sener2017active} or for topic summarization~\cite{bairi2015summarization}. In this paper, we present a framework (\modelss) of data subset selection for \emph{selecting} a subset of unlabelled examples for obtaining labels complementary to the labelling functions. 

%% file: 3.methodology.tex
\section{Methodology}
\subsection{Problem Description}
Let $\Xcal$ and $\Ycal \in \{1...K\}$ be the feature space and label space, respectively. We also have access to $m$ labelling functions (LF) $\lambda_1$ to $\lambda_m$. As mentioned in Section~\ref{sec:motivating}, each  LF $\lambda_j$ is designed to record some class; let us denote\footnote{We use the association of LF $\lambda_j$ with some class $k_j$ only in the quality guide component (QG) of the loss in eqn. \ref{eq:objective}} by $k_j \in \{1...K\}$, the class associated with $\lambda_j$. 
The dataset consists of 2 components, {\em viz.},\\ 
    \noindent 1. $\Lcal = \{(\bfx_1, y_1, \bfl_1), (\bfx_2, y_2, \bfl_2), \dots ,\\ (\bfx_N, y_N, \bfl_N)\}$, which denotes the labelled dataset and \\
    \noindent 2. $\Ucal$ = $\{(\bfx_{N+1}, \bfl_{N+1}), (\bfx_{N+2}, \bfl_{N+2}), \dots ,\\ (\bfx_M,\bfl_M)\}$, which denotes the unlabelled dataset wherein $\bfx_i \in \Xcal$ , $y_i \in \Ycal$. \\
Here, the vector $\bfl_{i} = (l_{i1}, l_{i2}, \dots , l_{im})$ denotes the firings of all the LFs on instance $\bfx_i$. Each $l_{ij}$ can be either $1$ or $0$. $l_{ij}=1$ indicates that the LF $\lambda_j$ has fired on the instance $i$
and $0$ indicates it has not. All the labelling functions are discrete; hence, no continuous scores are associated with them.

\subsection{Classification and Labelling Function Models}
\model \xspace has a feature-based classification model $f_\phi(\bfx)$ which takes the features as input and predicts the class label. Examples of $f_\phi(\bfx)$ we consider in this paper are logistic regression and neural network models. The output of this model is $P_\phi^f(y|\bfx)$, {\em i.e.}, the probability of the classes given the input features. This model can be a simple classification model such as a logistic regression model or a simple neural network model. 

We also use an LF-based graphical model $P_{\theta}(\bfl_i, y)$ which, as specified in equation (\ref{eq:joint1}) for an example $\bfx_i$, is a generative model on the LF outputs and  class label $y$. 
\begin{equation}
\displaystyle P_{\theta}(\bfl_i, y) = \frac{1}{Z_\theta} \prod_{j=1}^m \psi_\theta(l_{ij}, y)
\label{eq:joint1}
\end{equation}
\vspace{-0.5cm}
\begin{equation}
    \psi_{\theta}(l_{ij},y) = 
\begin{cases}
    \exp(\theta_{jy})  & \text{if $l_{ij}\ne 0$} \\
    1 & \text{otherwise.}
\end{cases}
\label{eq:decoupledthetas}
\end{equation}
  There are $K$ parameters $\theta_{j1},\theta_{j2}...\theta_{jK}$ for each LF $\lambda_j$, where $K$ is the number of classes. 
  The model makes the simple assumption that each LF $\lambda_j$ independently acts on an instance $\bfx_i$ to produce  outputs $l_{i1}, l_{1i}...l_{im}$. The potentials $\psi_\theta$ invoked in equation (\ref{eq:joint1}) are defined in equation (\ref{eq:decoupledthetas}).  $Z_\theta$ is the normalization factor. 
We  propose a joint learning algorithm with semi-supervision to employ both features and LF predictions in an end-to-end manner.

\subsection{Joint Learning in \model} 
\label{sec:spear}
We first specify the objective of \model \xspace and thereafter explain each of its components in greater detail:
\small{\begin{align}\nonumber
\min_{\theta, \phi} &\sum_{i \in \Lcal} L_{CE}\left(P_\phi^f(y|\bfx_i), y_i\right) + \sum_{i \in \Ucal} H\left(P_\phi^f(y|\bfx_i)\right)
\\ &+\sum_{i \in \Ucal} L_{CE}\left(P_\phi^f(y|\bfx_i), g(\bfl_i)\right) \nonumber
 + LL_s(\theta| \Lcal)  
 \\ &+ LL_u(\theta| \mathcal U) + \sum_{i \in \Ucal \cup \Lcal} KL\left( P_\phi^f(y|\bfx_i),P_\theta(y|\bfl_i)\right) \nonumber
 \\ & + R(\theta|\{q_j\})
\label{eq:objective}
\end{align}}
\normalsize

\begin{table}[]
\centering
\begin{adjustbox}{width=0.48\textwidth}
\begin{tabular}{|c|l|}
\hline
\textbf{Notation}      & \textbf{Description}                                             \\ \hline
$f_\phi$ & The feature-based Model                                           \\ \hline

$P_\phi^f$                      & The label probabilities as per the feature-based model $f_\phi$       \\ \hline

$P_\theta$               & The label probabilities as per the LF-based Graphical Model                             \\ \hline
$L_{CE}$                      & Cross Entropy Loss:      $L_{CE}\left(P_\phi^f(y|\bfx), \tilde{y}\right) = -\log\left(P_\phi^f(y=\tilde{y}|\bfx)\right)$                                \\ \hline
$H$                      & Entropy function :
$H(P_\phi^f(y|\bfx)) = -\sum \limits_{\hat{y}} P_\phi^f(y=\hat{y}|\bfx) \log P_\phi^f(y=\hat{y}|\bfx)$
\\ \hline
$g$                      & Label Prediction from the LF-based graphical model               \\ \hline
$LL_s$                  & Supervised negative log  likelihood                     \\ \hline
$LL_u$                  & Unsupervised negative log likelihood summed over labels \\ \hline
KL                     & KL Divergence between two probability models \\ \hline
$R$                      & Quality Guide based loss                               \\ \hline
\end{tabular}
\end{adjustbox}
\caption{Summary of notation used.}
\label{tab:notation}
\end{table}
\vspace{-2ex}
Before we proceed further, we refer the reader to Table~\ref{tab:notation} in which we summarise the notation built so far as well as the notation that we will soon be introducing. 

\noindent \textbf{First Component (L1): } The first component (L1) of the loss $L_{CE}\left(P_\phi^f(y|\bfx_i), y_i\right) = -\log\left(P_\phi^f(y=y_i|\bfx_i)\right)$ is the standard cross-entropy loss on the labelled dataset $\Lcal$ for the model $P_\phi^f$.  

\noindent \textbf{Second Component (L2): } The second component L2 is the semi-supervised loss on the unlabelled data $\Ucal$. In our framework, we can use any unsupervised loss function. However, for this paper, we  use the Entropy minimisation~\cite{grandvalet2005semi} approach. Thus, our second component $H\left(P_\phi^f(y|\bfx_i)\right)$ is the entropy of the predictions on the unlabelled dataset. It acts as a form of semi-supervision by trying to increase the confidence of the predictions made by the model on the unlabelled dataset. 

\noindent \textbf{Third Component (L3): } The third component $L_{CE}\left(P_\phi^f(y|\bfx_i), g(\bfl_i)\right)$ is the cross-entropy of the classification model using the hypothesised labels from CAGE ~\cite{oishik} on $\Ucal$. 
Given that $\bfl_i$ is the output vector of all labelling functions for any $\bfx_i \in \Ucal$, we specify the predicted label for $\bfx_i$ using the LF-based graphical model $P_\theta(\bfl_i, y)$ from eqn. (\ref{eq:joint1}) as: $g(\bfl_i) = \mbox{argmax}_y  P_\theta(\bfl_i, y)$\\
\noindent \textbf{Fourth Component (L4): } The fourth component $LL_s(\theta|\Lcal)$ is the (supervised) negative log likelihood loss on the labelled dataset $\Lcal$ as per eqn. (\ref{eq:objective}):  
    $LL_s(\theta|\Lcal) = - \sum \limits _{i=1}^{N} \log P_\theta(\bfl_i, y_i)$

\noindent \textbf{Fifth Component (L5): } The fifth component $LL_u(\theta|\Ucal)$ is the negative log likelihood loss for the unlabelled dataset $\Ucal$ as per eqn. (\ref{eq:objective}). Since the true label information is not  available, the probabilities need to be summed over $y$:  
    $LL_u(\theta|\Ucal) = - \sum \limits _{i=N+1}^{M} \log \sum \limits _{y \in \Ycal} P_\theta(\bfl_i, y)$

\noindent \textbf{Sixth Component (L6): } The sixth component $KL(P_\phi^f(y|\bfx_i), P_\theta(y|\bfl_i))$ 
is the Kullback-Leibler (KL) divergence between the predictions of both the models, {\em viz.}, feature-based model $f_\phi$ and the LF-based graphical model $P_\theta$  summed over every example $\bfx_i \in \Ucal \cup \Lcal$. Through this term, we try and make the models agree in their predictions over the union of the labelled and unlabelled datasets.

\noindent \textbf{Quality Guides (QG): } As the last component in our objective, 
we use quality guides $R(\theta|\{q_j\})$ on LFs, which have been shown in ~\cite{oishik} to stabilise the unsupervised likelihood training while using labelling functions. Let $q_j$ be  the fraction of cases where $\lambda_j$  correctly triggered, and let $q_j^t$ be the user's belief on the fraction of examples $\bfx_i$ where $y_i$ and $l_{ij}$ agree. If the user's beliefs were not available, we consider the precision of the LFs on the validation set as the user's beliefs. Except for the SMS dataset, we take the precision of the LFs on the validation set as the quality guides. If $P_\theta(y_i=k_j|l_{ij}=1)$ is the model-based precision over the LFs, the quality guide based loss  can be expressed as $R(\theta | \{q_j^t\}) = \sum_j  q_j^t \log P_\theta(y_i=k_j|l_{ij}=1) + (1-q_j^t)  \log (1-P_\theta(y_i=k_j|l_{ij}=1))$. Throughout the paper, we consider QG always in conjunction with Loss \textbf{L5}.

In summary, the first three components (L1, L2 and L3) invoke losses on the supervised model $f_{\phi}$. While L1 compares the output $f_{\phi}$ against the ground truth in the labelled set $\Lcal$, L2 and L3 operate on the unlabelled data $\Ucal$ by minimizing the entropy of $f_{\phi}$ (L2) and by calibrating the $f_{\phi}$ output against the noisy predictions $g(\bfl_i)$ of the graphical model $P_{\theta}(\bfl_i,y)$  for each  $\bfx_i \in \Ucal$ (L3). The next two components L4 and L5 focus on maximizing the likelihood of the parameters $\theta$ of $P_{\theta}(\bfl_i,y)$ over labelled $\bfx_i \in \Lcal$ and unlabelled $\bfx_i \in \Ucal$ datasets respectively. Finally, in L6, we compare the probabilistic outputs from the supervised model $f_{\phi}$ against those from the graphical model $P_{\theta}(\bfl,y)$ through a KL divergence based loss. We use the ADAM (stochastic gradient descent) optimizer to train the non-convex loss objective. 

Previous data programming approaches~\cite{bach2019snorkel,oishik} adopt a cascaded approach in which they first optimise a variant of L5 to learn the $\theta$ parameters associated with the LFs and thereafter use the noisily generated labels using $g(\bfl)$ to learn the supervised model $f_{\phi}$ using a variant of L3. In contrast, our approach learns the LF's $\theta$ parameters and the model's $\phi$ parameters jointly in the context of the unlabelled data $\Ucal$.

We present synthetic experiments to illustrate the effect of \model \xspace for data programming and semi-supervision in a controlled setting  in which (i) the overlap between classes in the data is controlled and (ii) the labelling functions are accurate. The details of the synthetic experiments are provided in the appendix. 

\subsection{\modelss: Subset Selection with \model}
\label{sec:spearss}
Suppose we are given an unlabelled data set $\Ucal$ and a limited budget for data labelling because of the costs involved in it. It is essential for us to choose the data points that need to be labelled properly. We explore two strategies for selecting a subset of data points from the unlabelled set. We then obtain the labels for this subset, and run \model\ on the combination of this labelled and unlabelled set. The two approaches given are intended to maximise diversity of the selected subset in the feature space.  We complement both the approaches with Entropy Filtering (also described below).

\noindent {\bf Unsupervised Facility Location:} In this approach, given an unlabelled data-set $\Ucal$, we want to select a subset $\Scal$ such that the selected subset has maximum diversity with respect to the features. Inherently, we are trying to maximise the information gained by a machine learning model when trained on the subset selected. The objective function for unsupervised facility location is  
$f_{\text{unsup}}(\Scal) = \sum_{i \in \Ucal} \max_{j \in \Scal} \sigma_{ij}$
where $\sigma_{ij}$ denotes the similarity score (in the feature space $\Xcal$) between data instance $\bfx_i$ in unlabelled set $\Ucal$ and data instance $\bfx_j$ in selected subset data $\Scal$. We employ a lazy greedy strategy to select the subset. In conjunction with Entropy Filtering described below,
we call this technique \emph{Unsupervised Subset Selection}. 

\noindent {\bf Supervised Facility Location:} The objective function for Supervised Facility Location~\cite{wei2015submodularity} is 
$f_{\text{sup}}(\Scal) = \sum_{y \in \Ycal} \sum_{i \in \Ucal_y} \max_{j \in \Scal \cap \Ucal_y} \sigma_{ij}$.
 Here we assume that $\Ucal_y \subseteq \Ucal$ is the subset of data points with hypothesised label $y$. Simply put, $\Ucal_y$ forms a partition of $\Ucal$ based on the hypothesized labels obtained by performing unsupervised learning with labelling functions. In conjunction with Entropy Filtering, we call this technique \emph{Supervised Subset Selection}.

\noindent \textbf{Entropy Filtering: } We also do a filtering based on entropy. In particular, we sort the examples based on maximum entropy and select $fB$ number of data points\footnote{In our experiments, we set $f = 5$}, where $B$ is the data selection budget (which was set to the size of the labelled set $|\Lcal|$ in all our experiments). On the filtered dataset, we perform the subset selection, using either the supervised or unsupervised facility location as described above. Below, we describe the optimisation algorithm for subset selection. 

\noindent {\bf Optimisation Algorithms and Submodularity: } Both $f_{\text{unsup}}(\Scal)$ and $f_{\text{sup}}(\Scal)$ are submodular functions. 
We select a subset $\Scal$ of the filtered unlabelled data, by maximising these functions under a cardinality budget $k$ ({\em i.e.}, a labelling budget). For cardinality constrained maximisation, a simple greedy algorithm provides a near-optimal solution \cite{nemhauser1978analysis}. Starting with $\Scal^0 = \emptyset$, we sequentially update 
\vspace{-1.5ex}
\begin{equation}
\Scal^{t+1} = \Scal^t \cup \displaystyle \underset{j \in \Ucal \backslash \Scal^t}{\mbox{argmax}} f(j | \Scal^t)
\label{eq:greedy}
\end{equation}

\vspace{-1ex}
where $f(j | \Scal) = f(\Scal \cup j) - f(\Scal)$ is the gain of adding element $j$ to set $\Scal$. We iteratively execute the greedy step (\ref{eq:greedy}) until $t = k$ and $|\Scal^t| = k$.
It is easy to see that the complexity of the greedy algorithm is $O(nkT_f)$, where $T_f$ is the complexity of evaluating the gain $f(j | \Scal)$ for the supervised and unsupervised facility location functions. We then significantly optimize this simple greedy algorithm  via a lazy greedy algorithm~\cite{minoux1978accelerated} via memoization and precompute statistics~\cite{iyer2019memoization}. 

%% file: 4.experiments.tex
\begin{table}[th!]
\centering
\begin{adjustbox}{width=0.48\textwidth}
\begin{tabular}{lcccccc} 
\toprule
Dataset & $|\mathcal{L}|$ & $|\mathcal{U}|$ & \#Rules/LFs & Precision & \%Cover  & \textbar{}Test\textbar{} \\ 
\toprule
MIT-R & 1842 & 64888 & 15 & 80.7 &14 & 14256 \\
YouTube & 100 & 1586 & 10 & 78.6 & 87& 250 \\
SMS & 69 & 4502 & 73 & 97.3 & 40 & 500 \\
Census & 83 & 10000 & 83 & 84.1 & 100 & 16281 \\
Ionosphere & 73 & 98 & 64 & 65 & 100 & 106 \\
Audit & 162 & 218 & 52 & 87.5 & 100 & 233 \\
IMDB & 284 & 852 & 25 & 80 & 58.1 & 500 \\
\bottomrule
\end{tabular}
\end{adjustbox}
\caption{Statistics of datasets and their rules/LFs. Precision refers to micro precision of rules. \%Cover is the fraction of instances in $\mathcal{U}$ covered by at least one LF. Size of Validation set is equal to $|\mathcal{L}|$.}
\label{tab:stats}
\end{table}



\section{Experiments}
In this section, we (1) evaluate our joint learning against state-of-the-art approaches and (2) demonstrate the importance of subset selection over random subset selection. 
 We present evaluations on seven datasets on tasks such as text classification, record classification and sequence labelling. 
 
\par
\subsection{Datasets}
We adopt the same experimental setting as in ~\citet{awasthi2020learning} for the dataset split and the labelling functions. However (for the sake of fairness), we set the validation data size to be equal to the size of the  labelled data-set unlike~\citet{awasthi2020learning} in which the size of the validation set was assumed to be much larger. 

We use the following datasets: \textbf{(1) YouTube}: A spam classification on YouTube comments; \textbf{(2) SMS Spam Classification~\cite{sms}}, which is a binary spam classification dataset containing 5574 documents; \textbf{(3) MIT-R \cite{mitr}}, is a sequence labelling task on each token with following labels: \textit{Amenity, Prices, Cuisine, Dish, Location, Hours, Others};  \textbf{(4) IMDB}, which is a plot summary based movie genre binary classification dataset, and the LFs (and the labelled set) are obtained from the approach followed by \citet{snuba}; \textbf{(5) Census}~\cite{census}, \textbf{(6) Ionosphere}, and \textbf{(7) Audit}, which are all UCI datasets. The task in the Census is to predict whether a person earns more than \$50K or not. Ionosphere is radar binary classification task  given a list of 32 features. The task in the Audit is to classify suspicious firms based on the present and historical risk factors.
    
Statistics pertaining to these datasets are presented in Table~\ref{tab:stats}. 
Since we compare performances against models that adopt different terminology, we refer 
to rules and labelling functions interchangeably.
For fairness, we restrict the size of the validation set and keep it equal to the size $|\Lcal|$ of the labelled set. For all experiments involving comparison with previous approaches, we used code and hyperparameters from~\cite{awasthi2020learning} but with our smaller-sized validation set. Note that we mostly outperform them even with their larger-sized validation set as can be seen in Table~\ref{tab:subset-main}. More details on training and validation set size are given in the appendix.
\par

\subsection{Baselines}
In Table~\ref{tab:subset-main}, we compare \model \xspace and \modelss \xspace against other following standard methods on seven datasets. \\
\textbf{Only-$\Lcal$}: We train the classifier $P_\theta(y|\bfx)$ only on the labelled data $\Lcal$ using loss component L1. As explained earlier, following~\cite{awasthi2020learning}, we observe that a 2-layered neural network trained with the small amount of labelled data is capable of achieving competitive  accuracy. We choose this method as a baseline and report gains over it. \\
{$\boldsymbol{\Lcal+\Ucal_{maj}}$}: We train the baseline classifier $P_\theta(y|x)$  on the labelled data $\Lcal$  along with  $\Ucal_{maj}$ where labels on the $\Ucal$ instances are obtained by majority voting on the rules/LFs.  The training loss is obtained by weighing instances labelled by rules as $\min_\theta \sum_{(\bfx_i, \bfl_i) \in \Lcal} -\log P_\theta(\bfl_i|\bfx_i) + \gamma \sum_{(\bfx_i,y_i)\in \Lcal} -\log P_\theta(y_i|\bfx_i)$. \\
\textbf{Learning to Reweight (L2R) \cite{l2r}}: This method trains the classifier by an online training algorithm that assigns importance to examples based on the gradient direction.\\
$\boldsymbol{\Lcal$+$\Ucal_{Snorkel}}$~\cite{ratner}: Snorkel's generative model that models class probabilities based on discrete LFs for consensus on the noisy and conflicting labels.\\
\textbf{Posterior Regularization (PR)~\cite{Hu2016HarnessingDN}}: This is a method for joint learning of a rule and feature network in a teacher-student setup. \\
\textbf{Imply Loss~\cite{awasthi2020learning}}: This approach uses additional information in the form of labelled rule exemplars and trains with denoised rule-label loss. Since it uses information in addition to what we assume,  Imply Loss can be considered as a skyline for our proposed approaches.

\begin{table*}[!h]
\centering
\begin{adjustbox}{width=1.0\textwidth}

\begin{tabular}{lccccccccc} 
\toprule
\multirow{2}{*}{Methods} & \multicolumn{7
}{c}{Datasets} \\ 
\cmidrule[\heavyrulewidth]{2-9}
  & \begin{tabular}[c]{@{}c@{}}YouTube\\$[$Accuracy$]$\end{tabular} & \begin{tabular}[c]{@{}c@{}}~~ SMS~ ~\\$[$F1$]$\end{tabular} & \begin{tabular}[c]{@{}c@{}}~ MIT-R~ ~\\$[$F1$]$\end{tabular} 
  & \begin{tabular}[c]{@{}c@{}}~IMDB~ ~\\$[$F1$]$\end{tabular} 
  & \begin{tabular}[c]{@{}c@{}}Census\\$[$Accuracy$]$\end{tabular}&
 \begin{tabular}[c]{@{}c@{}}~Ionosphere~\\$[$F1$]$\end{tabular}
 &
 \begin{tabular}[c]{@{}c@{}}~Audit (Imb)~\\$[$F1$]$\end{tabular}
 &
 \begin{tabular}[c]{@{}c@{}}~Audit (Bal)
 ~\\$[$F1$]$\end{tabular}
 \\ 
\toprule
Only-$\Lcal$ (Handpicked)& 90.7 {\scriptsize (1.2)} & 90.0 {\scriptsize (3.7)} & 74.1 {\scriptsize (0.4)} & 72.2 {\scriptsize (3.1)} & 78.3 {\scriptsize (0.3)} & 92.7 {\scriptsize (0.5)}&24.7{\scriptsize (2.6)} & 87.3 {\scriptsize (0.9)}\\ 
\hline\hline
$\Lcal$+$\Ucal_{maj}$  (Handpicked) & +1.9 {\scriptsize (1.1)} & -0.3 {\scriptsize (1.4)}& +0.1{\scriptsize (0.2)} & +1.2 {\scriptsize (0.3)}& -0.9 {\scriptsize (0.4)}& +0.4 {\scriptsize (0.7)}& -4.8 {\scriptsize (6)}& -1.4{\scriptsize (4.2)}\\
L2R~\cite{l2r}  (Handpicked)  & -3.7 {\scriptsize (5.1)} & +0.7 {\scriptsize (2.9)}& -20.2{\scriptsize (0.9)} & +4.5{\scriptsize (0.2)} &+3.6{\scriptsize (0.3)}& -18.8 {\scriptsize (0.3)} & -1.2{\scriptsize (3.2)} & -3.0{\scriptsize (4.9)}\\
$\Lcal$+$\Ucal_{Snorkel}$~\cite{ratner}  (Handpicked)  & +0.9 {\scriptsize (2.6)}& +0.3 {\scriptsize (4.5)}& -0.3{\scriptsize (0.2)} &+0.6{\scriptsize (1.8)} & +1.7 {\scriptsize (0.2)}& -0.6 {\scriptsize (0.5)}& -7.4{\scriptsize (3.4)} & -0.6{\scriptsize (4.2)}\\
Posterior Reg~\cite{Hu2016HarnessingDN} (Handpicked)  & -1.9{\scriptsize (1.6)} & -3.3 {\scriptsize (1.9)}& -0.2 {\scriptsize (0.2)}& +1.1 {\scriptsize (0.7)}& -1.9 {\scriptsize (0)} & -0.1{\scriptsize (0.7)} & - & +0.1 {\scriptsize (1.4)}\\
ImplyLoss~\cite{awasthi2020learning}  (Handpicked) & +0.4 {\scriptsize (0.5)}& +0.9{\scriptsize (0.9)}&0.9{\scriptsize (0.4)} & +4.3 {\scriptsize (1.5)} & +3.4 {\scriptsize (0.1)}& -3.9{\scriptsize (2.4)} & - & +0.5{\scriptsize (1)}\\
\textbf{\model \xspace (Handpicked)}  & +3.7{\scriptsize (0.5)} & \textbf{+3.4}{\scriptsize (0.9)} & -0.8 {\scriptsize (0.5)} & +4.9{\scriptsize (0.3)} &\textbf{ +3.7} {\scriptsize (0.3)}& \textbf{+5.4} {\scriptsize (0.3)} & +44 {\scriptsize (0.9)}& \textbf{+4.3} {\scriptsize (0.9)} \\
\hline \hline 
\textbf{\modelss \xspace (Random Subset Selection)} & +3.5 & +1.8 & -2.9 & +4.0 & -5.2 & +4.7 & +41.7 & +2.0 \\
\textbf{\modelss \xspace (Unsupervised Subset Selection)}  & +3.9 & +1.9 &  +2.6 & -0.6 & +2.5 & +4.8 & +43.5 & +3.3 \\
\textbf{\modelss \xspace (Supervised Subset Selection)} & \textbf{+4.2} & +3.2 & +\textbf{2.9} & \textbf{+6.3} & +2.5  & {+5.1} & \textbf{+44.5} & +3.5 \\
\bottomrule
\end{tabular}
\end{adjustbox}
\caption{Performance of \model \xspace and \modelss \xspace for three subset selection schemes on seven data-sets. All numbers reported are gains over the baseline method (Only-$\Lcal$). All results are averaged over 5 runs. Numbers in brackets `()' represent standard deviation of the original score. Handpicked instances refers to instances selected from the dataset for designing LFs. These instances are taken directly from \cite{awasthi2020learning} to ensure fair comparison. 
}
\label{tab:subset-main}
\end{table*}



\subsection{Results with \model}
\model \xspace uses the `best' combination of the loss components L1, L2, L3, L4, L5, L6. To determine the `best' combination, we perform a grid search over various combinations of losses using validation accuracy/f1-score as the criteria for selecting the most appropriate loss combination. Imply Loss uses a larger-sized validation set to tune their models. In our experiments, we maintained a validation set size equal to the size of the labelled data. 
In Table~\ref{tab:subset-main}, we observe that \model \xspace performs significantly better than all other approaches on all but the MIT-R data-set. Please note that all results are based on the same hand-picked labelled data subset as was chosen in prior work~\cite{awasthi2020learning,snuba}, except for Audit and Ionosphere. Even though we do not have rule-exemplar information in our model, \model \xspace achieves better gains than even ImplyLoss. Recall that the use of ImplyLoss can be viewed as a skyline approach owing to the additional exemplar information that associates labelling functions with specific labelled examples.  
The slightly lower performance of the `best' \model \xspace on the MIT-R data-set can be partially explained by the fact that there are no LFs corresponding to the `0' class label, owing to which our graphical model is not trained for all classes. 
However, as we will show in the next section, by suitably determining a subset of the data-set that can be labelled (using the facility location representation function), we achieve improved performance even on the MIT-R data-set (see Table~\ref{tab:subset-main}). Also, note that in Table~\ref{tab:subset-main}, we present results on two versions of Audit, one in which both the train and test set are balanced, and the other where the labelled training set is imbalanced. In the imbalanced case (where the number of positives are only $10$\%), we were unable to successfully run the ImplyLoss and Posterior-Reg models (and hence \xspace the `-'), despite communication with the authors. We see that \model \xspace and similarly, \modelss \xspace (discussed below) significantly outperform the baselines by almost $40$\% in the imbalanced case. In the balanced case, the gains are similar to what we observe on the other datasets.

\subsection{Results with \modelss\label{sec:subsetselection} }
 Recall that all results discussed so far (including those for \model) on the Youtube, SMS, MIT-R, IMDB and Census datasets were based on the same `hand-picked' labelled data subset as in prior work~\cite{awasthi2020learning,snuba}. In the case of Audit and Ionosphere, the labelled subset was randomly picked. 
 In Table~\ref{tab:subset-main}, we summarise the results obtained by employing supervised and unsupervised subset selection schemes for picking the labelled data-set and present comparisons against results obtained using (i) `hand-picked' labelled data-sets, and (ii) random selection of the labelled set. In each case, the size of the subset is the same, which we set to be the size of the hand-picked labelled set. Our data selection schemes are applied to the `best' \model\   model obtained across various loss components. 
We observe that the best-performing model for the supervised and unsupervised data selection tends to outperform the best model based on random selection. Secondly, we observe that between the supervised and unsupervised data selection approaches, the supervised one tends to perform the best, which means that using the hypothesised labels does help. Thirdly, we observe that YouTube, MIT-R, IMDB and Audit using the selected subset outperform prior work that employ hand picked data-set, 
whereas, in the case of SMS, Census and Ionosphere, we come close. Finally, our approach is more stable than other approaches as the standard deviation of \model \xspace is low for 5 different runs across all the datasets.  

As an illustration, the examples such as $S_3$ and $S_4$ referred to in Section~\ref{sec:motivating} were precisely obtained through supervised subset selection in \modelss, to form part of the labelled dataset. As previously observed in Table~\ref{tab:insufficiency}, $S_3$ and $S_4$ \emph{complement} (via n-grams features such as F1 and F2) the effect of the labelling functions LF1 and LF2 on the unlabelled examples such as $S_1$ and $S_2$, when included in the labelled set. Further detailed results with subset selection, {\em etc.} can be found in the appendix. 
In general, we observe that when the subset  of instances selected for labelling is \emph{complementary} to the labelling functions (as in our case), the performance is higher than when the labelled examples (exemplars) are inspired by labelling functions themselves as done in  the work by ~\citet{awasthi2020learning}.

\subsection{Significance Test}
We employ the Wilcoxon signed-rank test \cite{wilcoxon1992individual} to determine whether there is a significant difference between \model \xspace and Imply Loss (current state-of-the-art). Our null hypothesis is that there is not significant difference between \model \xspace and Imply loss. For $n=7$ instances, we observe that the one-tailed hypothesis is significant at $p < .05$, so we reject the null hypothesis. Clearly, \model \xspace significantly outperforms Imply loss and, therefore, all previous baselines. 

Similarly, we perform the significance test to assess the difference between \modelss \xspace and Imply Loss. As expected, the one-tailed hypothesis is significant at $p < 0.05$, which implies that our \modelss \xspace approach significantly outperforms Imply Loss, and thus all other approaches.

\begin{table}
\Large
\centering
\begin{adjustbox}{width=0.48\textwidth}
\begin{tabular}{p{0.4\textwidth}cccc}
\toprule
\multirow{2}{*}{Methods} & \multicolumn{4}{c}{Datasets   } \\ 
\cmidrule[\heavyrulewidth]{2-5}
 & \begin{tabular}[c]{@{}c@{}}YouTube\\$[$Accuracy$]$\end{tabular} & \begin{tabular}[c]{@{}c@{}}~ ~ SMS~ ~~\\$[$F1$]$\end{tabular} & \begin{tabular}[c]{@{}c@{}}~ ~ MIT-R~ ~ ~\\$[$F1$]$\end{tabular} & \begin{tabular}[c]{@{}c@{}}Census\\$[$Accuracy$]$\end{tabular} \\ 
\toprule
ImplyLoss~\cite{awasthi2020learning}  & 94.1 & 93.2 & 74.3 & 81.1\\\hline
\textbf{\model \xspace (Handpicked)}  & +0.3 & \textbf{+0.2} & -0.9 & \textbf{+0.9} \\ \hline
\textbf{\modelss \xspace (Supervised Subset Selection)}   & \textbf{+0.8} & 0.0 & \textbf{+1.7} & -0.3 \\
\hline 
\bottomrule
\end{tabular}
\end{adjustbox}
\caption{Comparison of \model \xspace and \modelss \xspace against ImplyLoss on subset of datasets from Table~\ref{tab:subset-main} for which ImplyLoss used a much larger validation set than $|\mathcal{L}|$. JL uses a validation set sizes equal to $|\mathcal{L}|$. }

\label{tab:main-compare-validation}
\end{table}


%% file: 5.conclusion.tex
\section{Conclusion}
We study how data programming can benefit from labelled data by learning a model (\model) that jointly optimises the consensus obtained from labelling functions in an unsupervised manner, along with semi-supervised loss functions designed in the feature space. We empirically  assess the performance of the different components of our joint loss function. As another contribution, we also study some subset selection approaches to guide the selection of the labelled subset of examples. We present the performance of our models and present insights on both synthetic and real datasets. While outperforming previous approaches, our approach is often better than an exemplar-based (skyline) approach that uses the additional information of the association of rules with specific labelled examples. 

%% file: appendix.tex




\section{Illustration of \model \xspace on a synthetic setting}
\begin{figure}[!ht]
\centering
\includegraphics[width=0.12\textwidth]{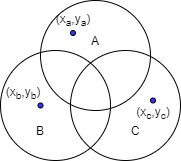}
\caption{Synthetic data}
\label{fig:synthetic}
\end{figure}

Through a synthetic example,
we illustrate the effectiveness of our formulation of combining semi-supervised learning with labelling functions ({\em i.e.}, combined Losses 1-6) to achieve superior performance. 
Consider a 3-class classification problem with overlap in the feature space as depicted in Figure \ref{fig:synthetic}. 
 The classes are A, B and C. Though we illustrate the synthetic setting in 2 dimensions, in reality, we performed similar experiments in three dimensions (and results were similar).
We randomly pick 5 points from each class $i \in \{a, b, c\}$, and corresponding to each such point $(x_i,y_i)$ we create a labelling function
based on its coordinates:
\begin{itemize}
    \item $LF_a$: Consider the point $(x_a,y_a)$. The corresponding LF will be: if $y \ge y_a$ return 1 (i.e. classify as class A) else return 0 (abstain). \item $LF_b$: Similarly for $(x_b, y_b)$ the LF  will return 1 if  $x \le x_b$ and else will return 0. 
    \item $LF_c$: The LF corresponding to $(x_c, y_c)$ will return 1 if $x \ge x_c$ and else will return 0.
\end{itemize}
These seemingly $15$ weak labelling functions (5 for each class) actually aid in classification when the labelled example set is extremely small and the classifier is unable to get a good estimate of the class boundaries. This can be observed in  Table~\ref{tab:synthetic-results} wherein we report the F1 score on a held out test dataset for models obtained by training on the different loss components. The results are reported in the case of three dimensions, wherein each circle was obtained as a 3-dimensional gaussian. The means for the three classes A, B and C were respectively, $(0,0,0)$, $(0,1,0)$ and $(0,0,1)$ and the variance for each class was set to  $(1, 1, 1)$. The  training and test sets had 1000 examples each, with roughly equal number of samples randomly generated from each class (gaussian).  In the first experiment (result in the first row of Table~\ref{tab:synthetic-results}), the training was performed on the  L1 loss by treating the entire data-set of 1000 examples as labelled. In all the other experiments only 1\% (10 examples with almost uniform distribution across the 3 classes) of the training set was considered to be labelled and the remaining (990  examples) were treated as unlabelled. 
\begin{table}[!h]
\centering
\begin{adjustbox}{width=0.48\textwidth}

\begin{tabular}{|l|l|}
\hline
Loss  component used for training & F1 Score \\ \hline
L1 (on entire dataset as labelled)       & 0.584    \\ \hline
L1 (1\% labelled) & 0.349    \\ \hline
L1 (1\% labelled) +L2  & 0.352    \\ \hline
L1 (1\% labelled)+L2+L3+L4 (1\% labelled)+L5+L6       & \textbf{0.440}    \\ \hline
L4 (1\% labelled)+L5 & 0.28 \\ \hline
\end{tabular}
\end{adjustbox}
\caption{F1 scores on test data in the synthetic setting}
\label{tab:synthetic-results}
\end{table}

We make the following important observations with respect to Table~\ref{tab:synthetic-results}:
        (1) {\bf Skyline:} When the entire training data is treated as labelled and loss function L1 is minimized, we obtain a skyline model with F1 score of $0.584$. 
        (2) With just $1\%$ labelled data on L1, 
        we achieve $0.349$ F1 score (using only the labelled data).
        (3) We obtain an F1 score of $0.28$ using  the labelling functions on the unlabelled data (for L5) in conjunction with the 1\% labelled data (for L4). 
        (4) When the $1\%$ labelled data (for L1) and the remaining observed unlabelled data (for L2) are used to train the semi-supervised model using L1+L2, an F1 score of $0.352$ is obtained.
        (5) However, by jointly learning on all the loss components, we  observe an F1 score of $0.44$. This is  far better than the numbers obtained using only (semi)-supervised learning  and those obtained using only the labelling functions. Understandably, this number is lower than the skyline of $0.584$ mentioned on the first row of Table~\ref{tab:synthetic-results}.

\section{Network Architecture}
\begin{table}[h]
\centering
\begin{adjustbox}{width=0.48\textwidth}
\begin{tabular}{lccc}
\hline
Datasets & lr (f network) & lr (g network) & Batch Size \\ \hline
SMS               & 0.0001                  & 0.01                    &    256                 \\
MIT-R             & 0.0003                  & 0.001                   &     512                \\
Census            & 0.0003                  & 0.001                   &    256                 \\
Youtube           & 0.0003                  & 0.001                   &      32               \\
Ionosphere        & 0.003                   &  0.01                       &        32             \\
Audit             &        0.0003               &     0.01                    &          32            \\
IMDB             &        0.0003               &     0.01                    &          32            \\\hline            
\end{tabular}
\end{adjustbox}
\caption{Hyper parameter details for the different datasets}
\label{tab:hyperparametes}
\end{table}

\begin{table*}
\centering
\begin{adjustbox}{width=0.7\textwidth}
\begin{tabular}{|c|c|c|c|c|c|} 
\hline
\multirow{2}{*}{Loss Combination} & \multicolumn{5}{c|}{Datasets} \\ 
\cline{2-6}
 & \begin{tabular}[c]{@{}c@{}}YouTube\\(Accuracy)\end{tabular} & \begin{tabular}[c]{@{}c@{}}~ ~ ~SMS~ ~ ~\\(F1)\end{tabular} & \begin{tabular}[c]{@{}c@{}}~ ~ MIT-R~ ~~\\(F1)\end{tabular}  & \begin{tabular}[c]{@{}c@{}}~ ~IMDB~ ~~\\(F1)\end{tabular}  & \begin{tabular}[c]{@{}c@{}}Census\\(Accuracy)\end{tabular}\\ 
\hline
L1+L2+L3+L4 & 94.6 & 93.1 & 72.5 & 73.6  & 82.0\\ 
\hline
L1+L2+L4+L6 & 92.0 & 91.9 & 69.7 &73.3  & 81.3 \\ 
\hline
L1+L3+L4+L6 & 94.4 & 93.2 & 29.8 & 74.4& 81.0\\ 
\hline
L1+L2+L3+L4+L6 & 94.4 & 92.3 & 29.5 & 64.4 & 80.9\\ 
\hline
L1+L3+L4+L5+L6  & \textbf{94.6} & \textbf{93.4} & \textbf{73.2} & \textbf{77.1} & \textbf{82.0}\\ 
\hline
L1+L2+L3+L4+L5+L6 & 94.5 & 93.0 & 72.8 & 76.9 & 81.9 \\
\hline
\end{tabular}
\end{adjustbox}
\caption{Performance on the test data, of various loss combinations from our objective function in equation (\ref{eq:objective}). For each dataset, the numbers in {\bf bold} refer to the `best' performing combination, determined based on performance on the validation data-set. In general, we observe that all the loss components (barring L2) contribute to the best model. Note that all combinations includes QG (Component 7).}
\label{tab:losscomponents}
\end{table*}

To train our model on the supervised data $L$, we use a neural network architecture having two hidden layers with ReLU activation. We chose our classification network to be the same as~\cite{awasthi2020learning}. In the case of MIT-R and SMS, the classification network contain 512 units in each hidden layer whereas the classification network for Census has 256 units in its hidden layers. For the YouTube dataset, we used a simple logistic regression as a classifier network, again as followed in~\cite{awasthi2020learning}. The features as well as the labelling functions for each dataset are also directly obtained from Snorkel \cite{ratner} and \cite{awasthi2020learning}. Please note that all experiments (barring those on subset selection) are based on the same hand-picked labelled data subset as was chosen in~\cite{awasthi2020learning}. \par
In each experiment, we train our model for 100 epochs and early stopping was performed based on the validation set. We use Adam optimizer with the dropout probability set to 0.8. The learning rate for $f$ and $g$ network are set to 0.0003 and 0.001 respectively for YouTube, Census and MIT-R datasets. For SMS dataset, learning rate is set to 0.0001 and 0.01 for $f$ and $g$ network. For Ionosphere dataset, learning rate for $f$ is set to 0.003. For each experiment, the numbers are obtained by averaging over five runs, each with a different random initialisation. The model with the best performance on the validation set was chosen for evaluation on the test set. As mentioned previously, the experimental setup in~\cite{awasthi2020learning} surprisingly employed a large validation set.
For fairness, we restrict the size of the validation set and keep it equal to the size of the labelled set. For all experiments involving comparison with previous approaches, we used code and hyperparameters from~\cite{awasthi2020learning} but with our smaller sized validation set. \par
Following~\cite{awasthi2020learning}, we used binary-F1 as an evaluation measure for the SMS, macro-F1 for MIT-R datasets, and accuracy for the YouTube and Census datasets.

\section{Optimisation Algorithms and Submodularity: Lazy Greedy and Memoization \label{sec:submodOpt}} 

Both $f_{\text{unsup}}(X)$ and $f_{\text{sup}}(X)$ are submodular functions, and for data selection, we select a subset $X$ of the unlabelled data, which maximises these functions under a cardinality budget (i.e. a labelling budget). For cardinality constrained maximisation, a simple greedy algorithm provides a near optimal solution~\cite{nemhauser1978analysis}. Starting with $X^0 = \emptyset$, we sequentially update $X^{t+1} = X^t \cup \mbox{argmax}_{j \in V \backslash X^t} f(j | X^t)$, where $f(j | X) = f(X \cup j) - f(X)$ is the gain of adding element $j$ to set $X$. We run this till $t = k$ and $|X^t| = k$, where $k$ is the budget constraint. It is easy to see that the complexity of the greedy algorithm is $O(nkT_f)$ where $T_f$ is the complexity of evaluating the gain $f(j | X)$ for the supervised and unsupervised facility location functions. This simple greedy algorithm can be significantly optimized via a lazy greedy algorithm~\cite{minoux1978accelerated}. The idea is that instead of recomputing $f(j | X^t), \forall j \notin ^t$, we maintain a priority queue of sorted gains $\rho(j), \forall j \in V$. Initially $\rho(j)$ is set to $f(j), \forall j \in V$. The algorithm selects an element $j \notin X^t$, if $\rho(j) \geq f(j | X^t)$, we add $j$ to $X^t$ (thanks to submodularity). If $\rho(j) \leq f(j | X^t)$, we update $\rho(j)$ to $f(j | X^t)$ and re-sort the priority queue. The complexity of this algorithm is roughly  $O(k n_R T_f)$, where $n_R$ is the average number of re-sorts in each iteration. Note that $n_R \leq n$, while in practice, it is a constant thus offering almost a factor $n$ speedup compared to the simple greedy algorithm. One of the parameters in the lazy greedy algorithms is $T_f$, which involves evaluating $f(X \cup j) - f(X)$. One option is to do a na\"{\i}ve implementation of computing $f(X \cup j)$ and then $f(X)$ and take the difference. However, due to the greedy nature of algorithms, we can use memoization and maintain a precompute statistics $p_f(X)$ at a set $X$, using which the gain can be evaluated much more efficiently~\cite{iyer2019memoization}. At every iteration, we evaluate $f(j | X)$ using $p_f(X)$, which we call $f(j | X, p_f)$.  We then update $p_f(X \cup j)$ after adding element $j$ to $X$. Both the supervised and unsupervised facility location functions admit precompute statistics thereby enabling further speedups. 

\section{Role of different components in the loss function}
Given that our loss function has seven components (including the quality guides), a natural question is `how do we choose among the different components for joint learning (JL)?' Another question we attempt to answer is `whether all the components are necessary for JL?' For our final model ({\em i.e.}, the results presented in Tables 6  and 7 of the main paper), we attempt to choose the best performing JL combination of the 7 loss components, {\em viz.} L1, L2, L3, L4, L5, L6. To choose the `best' JL combination, we evaluate the performance on the validation set of the different JL combinations. Since we generally observe considerably weaker performance by selecting lesser than 3 loss terms, we restrict ourselves to 3 or more loss terms in our search. We report performance on the test data, of various JL combinations from our objective function for each of the four data-sets. For each data-set, the numbers in {\bf bold} refer to the `best' performing JL combination, determined based on performance on the validation data-set. 

The observations on the results are as follows. 
Firstly, we observe that all the loss components (barring L2 for three datasets) contribute to the best model. 
Furthermore, we observe that the best JL combination  (picked on the basis of the validation set) either achieves the best performance or close to best among the different JL combinations as measured on the test dataset. Secondly, we observe that QGs do not cause significant improvement in the performance during training.
